\documentclass[sigconf]{acmart}
\renewcommand\footnotetextcopyrightpermission[1]{} 
\usepackage{url}
\usepackage{graphicx}  
\usepackage{wrapfig}  
\usepackage{multicol}
\usepackage{amsmath,amsfonts,amsthm,bm}
\usepackage[linesnumbered,ruled,vlined]{algorithm2e}

\newcommand{\eat}[1]{}
\newcommand{\system}{AITEST}

 \setcopyright{none}
\usepackage{wrapfig}
\usepackage{amsmath}
\usepackage{graphics}
\usepackage{graphicx}
\usepackage{latexsym}
\usepackage{array}
\usepackage{enumitem}
\newcolumntype{M}[1]{>{\centering\arraybackslash}m{#1}}
\usepackage{multirow}
\usepackage{url}
\usepackage{listings,color,colortbl}

\definecolor{light-gray}{gray}{0.80}

\usepackage{balance}

\begin{document}

\title{Automated Testing of AI Models}

\author{Swagatam Haldar, Deepak Vijaykeerthy, Diptikalyan Saha}
\email{swagatam.haldar@ibm.com,   deepakvij@in.ibm.com, diptsaha@in.ibm.com}
\affiliation{%
  \institution{IBM Research AI}
  \country{India}
}

\newcommand{\norm}[1]{\left\lVert#1\right\rVert}

\begin{abstract}
The last decade has seen tremendous progress in AI technology and applications. With such widespread adoption, ensuring the reliability of the AI models is crucial. In past, we took the first step of creating a testing framework called AITEST for metamorphic properties such as fairness, robustness properties for tabular, time-series, and text classification models. In this paper, we extend the capability of the AITEST tool to include the testing techniques for Image and Speech-to-text models along with interpretability testing for tabular models. These novel extensions make AITEST  a comprehensive framework for testing AI models.
 
\end{abstract}

\maketitle
\pagestyle{plain} 

 
\section{Introduction}
\label{sec:intro}

AI has been used in many diverse applications where the decision taken by the model directly impacts human life. It is therefore of utmost importance to make AI models as reliable as possible. Unfortunately, data scientists predominantly look for improving the accuracy or other generalization properties such as precision, recall, etc., often ignoring critical properties such as fairness, robustness while training the model. As a result, we have seen many instances of unfairness/bias and robustness issues in scenarios such as recidivism~\cite{skeem2007assessment}, job applications\cite{Recruitment}, etc. 

Even though there are many techniques present for testing~\cite{MLTEST}, there has been a dearth of comprehensive ML model testing tools which can work across modalities, different types of models, and go beyond generalizability. There exist few toolkits like AIF360~\cite{AIF360}, CHECKLIST~\cite{CHECKLIST} which either concentrates on a single property such as fairness or single modality such as text, but a comprehensive set of testing algorithms under a common framework is scarce.  

In~\cite{AITEST}, we presented a tool called \system{}, a framework for testing black-box models to solve the above problems. It has covered tabular, text, and time-series modalities and focused on generalizability, fairness, and robustness properties.  Yet, our coverage was incomplete when it comes to testing a variety of models used by our clients~\cite{IGNITE}. 

In this paper, we include a variety of testing algorithms in our tool in three modalities viz. tabular, image, speech-to-text. Extensible design of \system{} helped us include such analysis without much change in the framework. 

One of the major problems of AI models is interpretability. While the accuracy of AI models has increased over the years, the models have become more complex and un-interpretable. Interpretability testing deals with whether a given black-box model can be effectively simulated by a given interpretable model. This is essentially related to global explainability~\cite{TREPAN} and essential related to model auditing, especially in the finance industry who predominantly uses models built on tabular data. We include tree-simulatibility testing to check whether a given black-box model can be simulated by a decision-tree model having interpretable characteristics specified by the user.  
We further add Image testing capabilities - essentially adding various kinds of transformations under which the model should be robust i.e. not change its decision. Specifically, we present eleven generic image transformations (inverse, saturation, etc.) and a Bayesian optimization-based algorithm for the adversarial attack on the black-block model.  

Industrial conversation systems often use a speech-to-Text (STT) engine to convert human speech to text before the text is fed into the dialog system. There are many instances in which the failure of the dialog system is attributed to the STT. We, therefore, introduce the testing capability of fairness and robustness properties for STT models.  Our contributions are  summarized below:
\begin{itemize}[leftmargin=*]
    \item We present a tool called \system{} with functionalities related to a) interpretability testing of tabular AI models, b) comprehensive sets of image transformations for testing black-box image classifiers, c) a comprehensive set of audio transformations for testing fairness and robustness properties for speech-to-text models.
    
    \item We present a short evaluation to demonstrate the effectiveness of the above techniques.

\end{itemize}

Our tool is part of IBM’s Ignite quality platform~\cite{IGNITE} and is used in testing multiple industrial AI models.  
    
The rest of the paper is organized as follows. The next section presents the flow of the tool. Section~\ref{sec:algo} describes our testing algorithms. Section~\ref{sec:expt} presents the experimental results. Section~\ref{sec:related} presents the related work. We conclude in Section~\ref{sec:conclusion}.

\section{Flow}
\label{sec:system}

\begin{wrapfigure}{l}{3cm}
  \begin{center}
    \includegraphics[width=3cm, height=3.7cm]{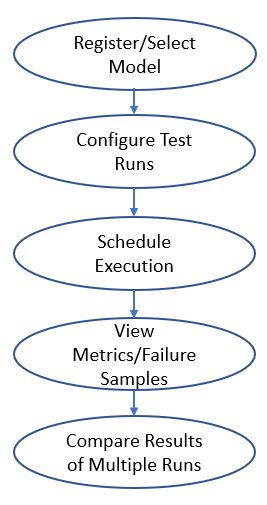}
  \end{center}
  \caption{Flow}
\end{wrapfigure}

\system{} uses a uniform flow across testing of models across different modalities (see left). The user (typically AI tester, data scientist or model risk manager) will start with registering a model into \system{} by providing model API details (endpoint, input template, output template, authentication headers), type of models (tabular-data classifier, time-series prediction, text-classifier, conversation-classifier, image-classifier, speech-to-text models), and seed data (training or test data which can be used in creating test samples). Users can select the newly registered model or an existing registered model and proceed to select the model type-specific properties and input necessary configurations required for each property (the threshold for fairness, transformation control parameters, etc.) to schedule a test run.  Users can view the status of the test run which shows the number of test cases generated and their status (passed/failed) specific to each selected property. Users can then view the metrics corresponding to the run for each completed property along with the failure samples and textual explanation. Users can compare the result of multiple runs.

\eat{
\begin{table}[t]
  \caption{Test Properties for different Model Types}
\centering
 \begin{tabular}{|l|l|p{0.25\linewidth}|}\hline`
Test Property & Metrics & Test Failure\\\hline
\multicolumn{3}{|c|}{Tabular data Classifier}\\\hline
Correctness & Accuracy & \multirow{4}{=}{Label mismatch with gold standard label}\\
  & Precision & \\
  & Recall & \\
  & F-Score & \\\hline
Group Discrimination (R) & Disparate Impact (DI) & \multirow{2}{=}{If DI $\notin R$}\\
  & Demographic Parity & \\\hline
Individual Discrimination & \multirow{2}{*}{Flip-rate} & \multirow{2}{=}{Label mismatch of the two samples}\\\cline{1-1}
Adversarial Robustness &  & \\\hline
\multicolumn{3}{|c|}{Text Classifier}\\\hline
Typo Sensitivity(L) & \multirow{6}{*}{Flip-rate} & \multirow{6}{=}{Label mismatch of original and transformed sample} \\\cline{1-1}
Noise Sensitivity(L) & & \\\cline{1-1}
Adjective Sensitivity & & \\\cline{1-1}
Tense Sensitivity & & \\\cline{1-1}
Voice Sensitivity & & \\\cline{1-1}
Paraphrase variation & & \\\hline
\multicolumn{3}{|c|}{Time Series Forecasting}\\\hline
Small Linear Change ($\alpha$) &  \multirow{3}{*}{RMSE change ($\Delta R$)} & \multirow{2}{=}{If $\Delta R > \alpha$}\\\cline{1-1}
Un-ordered data ($\alpha$) & &\\\cline{1-1}\cline{3-3}
Large Linear Change ($\beta$) &  & If $\Delta R < \beta$ \\\hline
  \end{tabular}
  \label{tab:1}
\end{table}

}
\section{Testing Algorithms}
\label{sec:algo}

\subsection{Interpretability Testing of Tabular Classifiers}
Understanding the decision making process of most machine learning models are hard for human beings as typically they are optimized over performance metrics and not designed to be interpretable.
The global explainability problem refers to creating an interpretable model which can mimic the reasoning of the target model for a given set of samples, usually the entire traiing data. Rule based models like decision trees or rule sets~\cite{breiman1984classification, lakkaraju2016interpretable, wang2017bayesian} and sparse linear models~\cite{LIME} are popular choices for such surrogates.

In our framework, we assess the \textit{interpretability} of a given target blackbox classification model $M$ by first constructing a decision tree~\cite{breiman1984classification} surrogate $T$ over the entire training data but with the predictions from the model instead of the training data labels. We then evaluate the following three decision tree interpretability characteristics that the user provides. Note that huge decision trees are not really interpretable. While building $T$, we leverage data generation~\cite{TREPAN} to overcome sample scarcity to decide \textit{attribute splits} at nodes far from the root. \eat{However, instead of random generation we use realistic data synthesis which better reflects the original data distribution in the \textit{query} samples.}

\paragraph{\textbf{Average Path Length (APL)}} It is defined~\cite{TREE-REGULARIZATION} as the expected no. of nodes along any root to leaf decision path or equivalently, the expected length of a path in $T$ which simulates the target model $M$. The expectation is computed over all such paths in the surrogate.
\begin{equation}
\textrm{APL} = \frac{\sum_{path \in T}\text{No. of decision nodes in } path}{\text{Total no. of }path\text{s in }T}
\end{equation}
Intuitively, this tries to measure the expected number of \textit{decisions} to be made to arrive at an outcome for a sample according to the model $M$. Each decision here is essentially a boolean predicate evaluation for some attributes of the sample.

\paragraph{\textbf{Maximum Path Length (MPL)}} It is the maximum number of nodes present in any decision path in the surrogate tree $T$.
\begin{equation}
\textrm{MPL} = {\max}_{path \in T} \text{ No. of decision nodes in }path
\end{equation}
This metric captures, in the worst case, how many decisions are to be made to predict the outcome of a sample. Interpretability favours \textit{small} values of APL and MPL.

\paragraph{\textbf{Fidelity}} It is computed as the percentage of test samples (from a reserved suite $\mathcal{D}$) for which predictions for labels from target model $M$ and the surrogate tree $T$ match i.e., are same.
\begin{equation}
\textrm{Fidelity} = \frac{\sum_{x \in \mathcal{D}} \mathbf{1}_{M(x) = T(x)}}{\text{Total no. of samples in $\mathcal{D}$}}
\end{equation}
where $x$ is a sample from $\mathcal{D}$.

Fidelity is a measure of how accurately the surrogate tree $T$ resembles the target model $M$ in terms of output and decision boundary.

The successful test of tree-simulatability testing generates a decision tree model which satisfies all the above characteristics.

\subsection{Image Classifier Testing}
Deep Neural Networks (DNNs) have found widespread usage in several Computer Vision (CV) tasks. Even though they match (or exceed) the performance of humans on these tasks, in practice, DNNs are vulnerable to malicious inputs~\cite{DEEPTEST, fgsm}. Consequently, there is an increasing need to develop techniques to test these models before deployment. In our framework, we propose a rich set of transforms (refer to table ~\ref{table:image}) to test DNNs for CV tasks, each of which checks whether the prediction of the model changes on the transformed images. 

The transforms used by our framework broadly fall into two categories. Static transforms such as translation, scaling, rotation, blurring, etc.~\cite{distortions}, and dynamic ones that generates black-box adversarial examples via Bayesian Optimisation~\cite{HARDLABELBBOX}. Static transforms can be further classified into the following three categories: linear, affine, and convolution.

\paragraph{Static Transforms} In the case of \textit{linear transforms}, we add, subtract or multiply either a constant or random noise to each pixel of an image. For instance, to change the brightness of a given image, we add (or subtract) a constant value, and in the case of random Gaussian noise transform, we add random noise sampled from a unit Gaussian distribution to the image. \textit{Affine transforms} modify the geometric structure of the image while preserving proportions of lines, but not necessarily the lengths and angles. Typically, affine transforms can be represented by a 2D matrix, and hence are amenable to compositions. In practice, these transforms (or their compositions) are used to simulate several image deformations which a model would encounter during a real-world deployment. \textit{Convolutions} are general purpose filters that work by multiplying a pixel's and its neighboring pixel's value by a matrix (aka a kernel matrix). Intuitively, the value of each transformed pixel is computed by adding the products of each surrounding pixel value with the corresponding kernel value. Transforms such as Blur, Fog, etc., are a few examples of convolution transforms.

\begin{table}
\centering
\footnotesize
\begin{tabular}{llp{0.4\linewidth}}
\hline
Name & Type & Description \\
\hline
Inverse & Linear & Inverts the image \\
Scale & Affine & Scales the image by a factor \\
Rotate & Affine & Rotates the image by an angle \\
Shear & Affine & Slants an image by an angle \\
Saturation & Linear & Perturbs the saturation levels of the images \\
Brightness & Linear & Changes the brightness of the image\\
Contrast & Linear & Changes the contrast of the image \\
Fog & Convolution & Simulates realistic fog on the image \\
Gaussian Blur & Convolution & Blurs the image through a random Gaussian Kernel \\
Zoom \& Blur & Convolution & Zooms to  part of the image \& blurs it \\
Gaussian Noise & Linear & Adds a random noise sampled from a unit Gaussian distribution to the image\\
Bayesian Optimisation & Adversarial & Generates adversarial examples via Bayesian Optimisation\\
\hline
\end{tabular}
\caption{Image transforms}
\label{table:image}
\end{table}

\paragraph{Dynamic Transforms} Adversarial examples are malicious inputs crafted by adversaries to fool DNNs. In our framework, we generate adversarial examples in a black-box setup where an adversary can only query the model via a predictive interface $f(x)$ and doesn't know any other information about the deployed model $f$. The adversarial noise added to an image is generated by solving the following constrained optimisation problem~\cite{HARDLABELBBOX}, where we aim to find an adversarial input $\mathbf{\hat{x}}_i$, which is close to $\mathbf{x}_i$, such that model's prediction changes.
\begin{equation}
  \begin{split}
    {\arg\max}_{y \in \{1, 2, \ldots, C\}}f(\mathbf{x}_i + \mathbf{\delta}) \neq y_i & \\ \mathbf{s.t.} \norm{\delta}_{\infty} \leq \delta_{max} &
\end{split}
\end{equation}
We solve the above problem via Bayesian Optimisation which offers an efficient approach to solve global optimisation problems. Typically, Bayesian Optimisers has 2 key ingredients, a surrogate model like a Gaussian process (GP) or a Bayesian Neural Network (BNN), and an acquisition function $\alpha(.)$ which provides us the next location to query the target function. Intuitively, an acquisition function balances the exploration and exploitation by giving higher value to the parts of the input space where the value of the target function is typically high and the surrogate model is very uncertain. The approach of using Bayesian Optimisation with a GP surrogate to attack the target model is described in Algorithm~\ref{alg:boalg}.

\begin{algorithm}
        \caption{Bayesian Optimisation Attack}
        \label{alg:boalg}
\begin{flushleft}
        \textbf{INPUT:} A black box function $f(x)$, seed data $\mathcal{D}_0$, no. of iterations T\\
        \textbf{OUTPUT:} Adversarial noise $\delta_T$
\end{flushleft}
\For{m = 1, 2, \ldots, T}{
Select $\delta_m = \arg\max \alpha_m(\delta | \mathcal{D}_{m - 1})$ \\
$y_m = f(x + \delta_m)$ and $\mathcal{D}_m \longleftarrow \mathcal{D}_{m - 1} \cup \{\delta_m, y_m\}$ \\
Update the surrogate model with $\mathcal{D}_m$
}
\end{algorithm}

\subsection{Speech-to-Text Model Testing}
Automatic speech recognition engines are being widely used in many automated tasks (voice assistants to a myriad of downstream activities) as such systems make it very convenient for humans to clarify their intents. Examples include \textit{Speech-to-Text} services from Google, Microsoft, Apple, Amazon, IBM etc. and also offline engines like Mozilla's \textit{DeepSpeech}~\cite{DEEPSPEECH}. The performance of the downstream tasks often depends on the precision of text transcription of the audio since it is fed to language understanding models next.

It is important to realize that humans are not always in a conducive surrounding for giving clear voice commands and regularly \textit{noise artefacts} get blended in the input speech. In our framework we test the \textit{robustness} of the  blackbox audio transcription models under white background noise and various interfering environmental perturbations to its input clips.
We also test for simple \textit{fairness} properties such as changing the gender or accent of the input speech having identical textual content.

We measure the \textit{Word Error Rates} (defined later) of text transcriptions to measure the effectiveness of our testing.

For the following paragraphs, we consider an audio clip as a one-dimensional signal of appropriately \textit{sampled} and \textit{normalized} values and denote the original clip by $\mathbf{x}$ and its corresponding perturbed clip by $\mathbf{x'}$.

\paragraph{White Noise Transform} Here we perturb the original clip by adding a random standard normal perturbation to each of its elements.

\begin{equation}
\mathbf{x'} = \mathbf{x} + \bm{\eta} 
\end{equation}
where each element of the sequence \bm{$\eta$} is sampled from a standard normal distribution $\mathcal{N}(0, \sigma^2)$.

The variance of the noise is decided by a given $SNR_{dB}$ (Signal to Noise Ratio) value that also specifies the loudness of the noise. $SNR_{dB}$ is a logarithmic scale measure of the ratio of \textit{power} of two signals which is related to their $RMS$ values as follows.
\begin{equation}
SNR_{dB} = 10\log \left(\frac{P_{signal}}{P_{noise}}\right) = 20\log \left(\frac{RMS_{signal}}{RMS_{noise}}\right)
\end{equation}
Also note that in case of standard normal distribution, since it has zero mean, its variance is the $RMS_{noise}$ value squared.
\begin{equation}
\sigma^2 = RMS_{noise}^2
\end{equation}
Using the above equations we can compute the perturbed signals for any desired $SNR_{dB}$ specified during testing.

\paragraph{Interference Transform} In this transformation, we overlay environmental noises that simulate different practical scenarios on top of the original clip. Example scenarios include restaurant, rainfall, festival, water dripping, windy noise etc. We control the signal strength by using the parameter $\theta$ while generating the linear combination of signal and noise that produces the interference effect.
\begin{equation}
\mathbf{x'} = \theta\mathbf{x} + (1-\theta)\bm{\eta} \text{ where } 0\leq\theta\leq1.
\end{equation}

\paragraph{Fairness Transforms} \textit{Speech-to-Text} transcription models are expected to have similar performance irrespective of the sensitive or protected attributes of the speaker. In this pretext, we test the models by synthesizing the perturbed input $\mathbf{x'}$ with the same spoken words or tokens in the same language but flipping its \textit{gender} (male to female and vice versa) or by changing the speaker's \textit{accent} such as Indian, American, French etc.

We synthesize the perturbed inputs using open-source speech generation engines or pretrained \textit{Text-to-Speech} models (such as \textit{Tacotron2}~\cite{TACOTRON} and \textit{Waveglow}~\cite{waveglow}) that synthesize audio from natural language transcripts directly. By training these models on audio from different speakers, we can leverage the learnt transformations.

These transforms try to capture the fairness of the models to different genders and varying ethnicity of the end-users who may be consuming the product or using in a critical decision making scenario. 

\section{Evaluation}

In this section we present some basic evaluation of our techniques of the key underlying novel testing techniques. This demonstrates that \system{} can be effectively used. 

\label{sec:expt}
\paragraph{Interpretability Testing for Tabular Data} In Table~\ref{table:interpret} we provide the interpretability metrics for some well known datasets: Adult Census Income~\cite{adult_data}, Bank Marketing~\cite{bank_data} and US Execution~\cite{execution_data} all obtained from the UCI repository~\cite{uci_repository}. We have taken blackbox tabular models from \textit{IBM Watson WML} framework. Here the user may set appropriate thresholds for the three metrics to decide whether the test fails or not.

\begin{table}
\centering
\footnotesize
\begin{tabular}{cccc}
\hline
Dataset & APL & MPL & Fidelity(\%) \\
\hline
Adult Income & 11.4 & 18 & 91.12 \\
Bank Marketing & 11.29 & 18 & 94.25 \\
Execution & 8.07 & 12 & 97.92 \\
\hline
\end{tabular}
\caption{Interpretability results}
\label{table:interpret}
\end{table}

\begin{figure}
    \includegraphics[width=7cm]{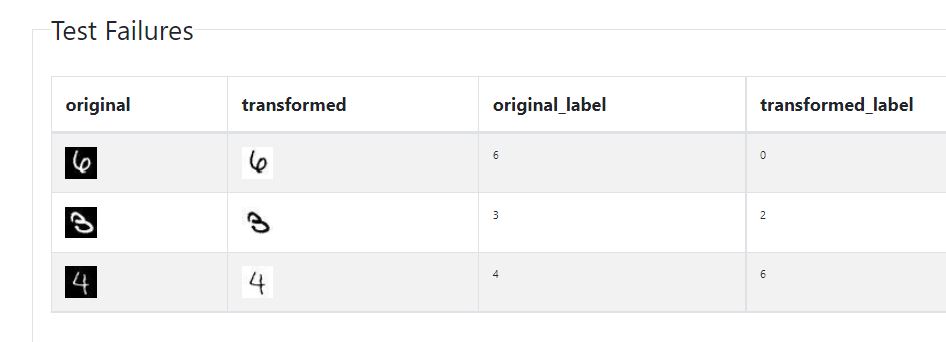}
  \caption{Failure Image  Transformations}
  \label{fig:image}
\end{figure}

\paragraph{Image}
To illustrate the effectiveness of our framework, we tested a couple of standard models on popular data sets LeNet5~\cite{lecun1998gradient} on MNIST~\cite{mnist1998lecun} and ResNet32~\cite{he2016deep} on CIFAR10~\cite{krizhevsky09learningmultiple}. From Table~\ref{table:image_result}, we can see that test cases generated by various transforms in our framework is quite effective in fooling the models. User can view such cases in the tool as shown in Figure~\ref{fig:image}.

\begin{table}
\centering
\footnotesize
\begin{tabular}{lrr}
\hline
Transform& \multicolumn{2}{c}{Dataset}\\
& MNIST & CIFAR10\\
\hline
None & 99.8 & 92.1 \\
Inverse & 97.1 & 86.1 \\
Scale & 96.1 &  88.2 \\
Rotate & 93.7 &  91\\
Shear & 97.8 & 85.1 \\
Saturation & 97.1 & 90.1 \\
Brightness & 97.3 & 90.3\\
Contrast & 96.5 & 90.5 \\
Fog & 91.7 & 86.4 \\
Gaussian Blur & 94 & 85.6 \\
Zoom \& Blur & 91 & 82.1 \\
Gaussian Noise & 92 & 89.4\\
Bayesian Optimisation & 10 & 9\\
\hline
\end{tabular}
\caption{Accuracy (\%) of various models on transformed example}
\label{table:image_result}
\end{table}

\paragraph{Speech-to-Text} For evaluating the perturbations, we use \textit{Word Error Rate (WER)} as a metric which is defined as the fraction of words omitted, substituted or newly inserted in the text transcription of the perturbed clip, when compared with the original text. If we have multiple audio samples, we simply the average the metric over all samples.

\begin{figure}
    \includegraphics[width=7cm]{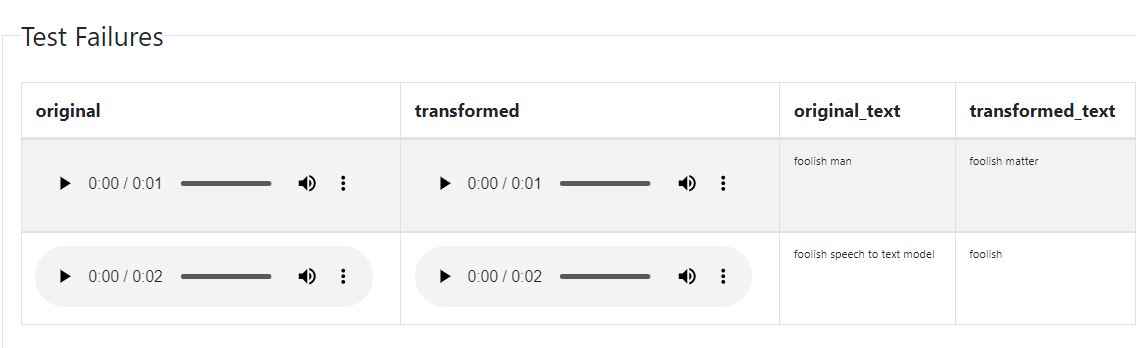}
  \caption{Speech-to-text Screenshots}
  \label{fig:s2t}
\end{figure}

We have tested it in a client-scenario which requires a voice assistant to process simple commands or queries. We provide some example failures detected by our method in Table~\ref{table:s2t}. User can play the original and transformed audio for the failure cases as shown in Figure~\ref{fig:s2t}. We direct interested readers to~\cite{AITEST_blogpost} and our supplement video that contain more examples and details.

\begin{table}
\footnotesize
\centering
\begin{tabular}{llll}
\hline
Transform type & Original Text & Perturbed Text & WER \\
\hline
White noise & \textit{I am ready} & \textit{find great if} & 1.0 \\
Restaurant noise & \textit{can I talk to someone} & \textit{can} & 0.8 \\
Water drip noise & \textit{keep holding} & \textit{keep clothing} & 0.5 \\
Wind noise & \textit{I need a minute} & \textit{I need a man} & 0.25 \\
French accent & \textit{repeat please} & \textit{he Peter please} & 1.0 \\

\hline
\end{tabular}
\caption{Speech-to-Text failure examples}
\label{table:s2t}
\end{table}

\eat{

\begin{figure}[b]
  \begin{tabular}{lr}
    \includegraphics[width=5cm]{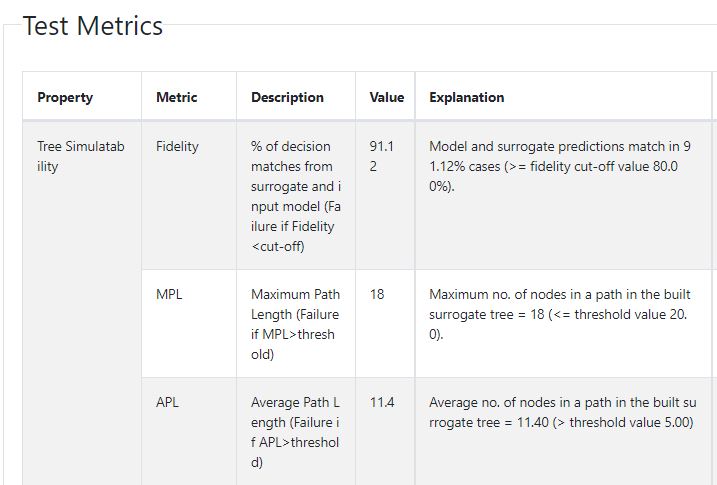}\\    \includegraphics[width=5cm, height=2cm]{images/Image.JPG}\\
    \includegraphics[width=5cm]{images/S2T.JPG}
  \end{tabular}
  \caption{\system{} Screenshots}
\end{figure}
}

\emph{A short video demonstration of our tool with a subset of properties include in this paper is available at \url{https://youtu.be/sc0YEHx4m0c}.}

\section{Related Work}
\label{sec:related}

In the recent past, there has been a lot of interest in developing techniques for testing DNNs for Computer Vision tasks that vary from simple transforms such a translation, rotation, etc.~\cite{DEEPTEST}, to richer formulations based on GANs~\cite{DEEPROAD}, Concolic Testing~\cite{DEEPCONCOLIC} \& adversarial examples generation~\cite{HARDLABELBBOX, fgsm}, we leverage some of these techniques in our framework. Similar techniques have also gained traction for speech recognition models. ~\cite{carlini2018audio} uses fully white-box gradient-based methods to generate targeted and untargeted attacks for models such as ~\cite{DEEPSPEECH}. Although the notion of simulatability and tree-simulatability are not new~\cite{TREE-REGULARIZATION} and testing a model for tree-simulatability or interpretability seems to be novel. Overall, \system{} remains the comprehensive tool for testing these many modalities.

\section{Conclusion and Future Work}
\label{sec:conclusion}

The last decade has seen various types of models and their applications. Only in recent time, efforts has been made to define a concrete Data and AI lifecycle~\cite{AILifecycle} and testing plays an important role in the lifecycle to ensure the reliability of AI models. We present a tool \system{} which encompasses variety of testing techniques across multiple modalities. In this paper, we include some novel capabilities such as interpretability testing and fairness testing of Speech-to-text models along with the implementation of some known properties. However, implementation of all the properties under one framework which provides easy usability is very useful for serving variety of client models.   

In future, we plan to add support for testing videos, multi-modal inputs, model compositions. The current implementation only supports black-box testing, which when configured, can be applied to a large number of other similar models of the same type without much change. We plan to add support for white-box testing algorithms.

\bibliographystyle{plain}
\bibliography{bib}

\end{document}